\documentclass[manuscript,screen, nonacm]{acmart}
\AtBeginDocument{%
  \providecommand\BibTeX{{%
    \normalfont B\kern-0.5em{\scshape i\kern-0.25em b}\kern-0.8em\TeX}}}

\begin{document}

\title[Virtual, Augmented, and Mixed Reality for Human-Robot Interaction]{Virtual, Augmented, and Mixed Reality for Human-Robot Interaction: A Survey and Virtual Design Element Taxonomy}

\author{Michael Walker}
\email{michael.e.walker@unc.edu}
\affiliation{%
  \institution{University of North Carolina at Chapel Hill}
  \city{Chapel Hill}
  \state{North Carolina}
  \country{United States}
}

\author{Thao Phung}
\affiliation{%
  \institution{Colorado School of Mines}
  \city{Golden}
  \state{Colorado}
  \country{United States}}
\email{thaophung@mines.edu}

\author{Tathagata Chakraborti}
\affiliation{%
 \institution{IBM Research AI}
 \city{Cambridge}
 \state{Massachusetts}
 \country{United States}}
\email{tchakra2@ibm.com}

\author{Tom Williams}
\affiliation{%
  \institution{Colorado School of Mines}
  \city{Golden}
  \state{Colorado}
  \country{United States}}
\email{twilliams@mines.edu}

\author{Daniel Szafir}
\affiliation{%
  \institution{University of North Carolina at Chapel Hill}
  \city{Chapel Hill}
  \state{North Carolina}
  \country{United States}}
\email{daniel.szafir@cs.unc.edu}


\begin{abstract}
Virtual, Augmented, and Mixed Reality for Human-Robot Interaction (VAM-HRI) has been gaining considerable attention in HRI research in recent years. However, the HRI community lacks a set of shared terminology and framework for characterizing aspects of mixed reality interfaces, presenting serious problems for future research. Therefore, it is important to have a common set of terms and concepts that can be used to precisely describe and organize the diverse array of work being done within the field. In this paper, we present a novel taxonomic framework for different types of VAM-HRI interfaces, composed of four main categories of virtual design elements (VDEs). We present and justify our taxonomy and explain how its elements have been developed over the last 30 years as well as the current directions VAM-HRI is headed in the coming decade.
\end{abstract}




\maketitle

\section{Virtual, Augmented, and Mixed Reality for Human-Robot Interaction}

Although robots are poised to increasingly support human society across a multitude of critical industries (e.g., healthcare, manufacturing, space exploration, agricultural) robot usage has remained limited due to the difficulties of robot control, supervision, and collaboration. A major source of this limitation arises from the bi-directional challenge of human-robot communication. Robots are often found to be incomprehensible, and humans struggle to predict robot capabilities or intentions. Simultaneously, robots lack the ability to reason about complex human behaviors: a skill inherently required for effective collaboration with humans.

At the heart of this problem lies the issue of poor information exchange between humans and robots, where neither can understand what the other is explicitly or implicitly conveying. This is analogous to the \textit{Gulf of Execution} and \textit{Gulf of Evaluation} concepts within the \textit{Human Action Cycle}, a proposed model describing human interactions with complex systems from the cognitive engineering and human-computer interaction communities \cite{norman1986user}. Humans regularly have issues conveying their high-level goals as inputs a robot can understand (e.g., gulf of execution), while robots often provide ineffective or no feedback to allow humans to assess the robotic system’s state (e.g., gulf of evaluation).

An example gulf of evaluation found in human-robot interactions is the motion inference problem, where robot users find the task of predicting when, where, and how a robot teammate will move to be difficult due to lack of information communicated from the robot. Information regarding a robot's planned movement is often invisible to human users, and even in circumstances where a robot is made to communicate its movement intent, the robot may lack the ability to share its motion plans as a human teammate would. A large amount of work in human-robot interaction (HRI) has looked at addressing this gulf of evaluation, such as by having robots use human-inspired social cues (e.g., gaze, gestures, etc.) to communicate their intentions \cite{sanghvi2011automatic}, altering robot trajectories to be more legible or expressive \cite{dragan2013legibility, szafir2014communication}, or using various other means such as auditory indicators \cite{tellex2014asking}. Although such techniques have shown effectiveness in reducing this gulf of evaluation by making robot motion more predictable, common constraints (e.g., computational, platform, environmental) may limit their feasibility when deployed in the real world. For example, an aerial robot’s morphology would prevent it from performing hand gestures or gaze, a dynamic or cluttered environment may restrict robots from altering from optimal trajectories to be more legible, and robot auditory indicators may be rendered ineffective if deployed by a noisy robot (e.g., aerial robots) or in a noisy environment.

To mitigate these issues, new methods of human-robot communication are being explored by HRI researchers that leverage more than the verbal or non-verbal cues seen in traditional human communication. New forms of visual communication have shown great promise in enhancing human-robot interaction, ranging from enhanced graphical displays to improve robot control that reduce gulfs of execution to LED lights that communicate various robot signals \cite{szafir2015communicating} that reduce gulfs of evaluation. Recently, the rise of consumer-grade, standardized virtual, augmented, and mixed reality (VAM) technologies (including the iPad, Microsoft HoloLens, Meta 2, Magic Leap, Oculus Rift/Quest, HTC Vive, etc.) has created a promising new medium for information exchange between users and robots and is well suited to enhance human-robot interactions in a variety of ways. VAM interfaces allow users to see 3D virtual imagery in a virtual space or contextually embedded within their environment. Up to this point, robot users have been forced to use traditional 2D screens to analyze the rich 3D data a robot often collects about its environment. VAM technology can also be used hands-free when in the form of a head-mounted display (HMD) that allows for more fluid and natural interactions with robots in a shared physical environment. Users can also be immersed in purely virtual worlds and interact with virtual robots which allows HRI researchers to evaluate interactions that would otherwise be impossible to observe either due to safety concerns or lack of access to an expensive physical robot(s). Finally, VAM interfaces allow HRI researchers to record and analyze human-robot interactions unlike ever before by leveraging the body, head, and gaze tracking inherent with VAM HMDs. 

This paper traces the development of early work merging HRI and VAM technology (which, while promising, was often hampered by limitations in underlying VAM technologies) and highlights more recent work that leverages modern systems. While this interdisciplinary surge of research is exciting and valuable, in research fields can raise disciplinary challenges. For example, many researchers are likely (unbeknownst to each other) to be simultaneously working on similar problems or using similar techniques, while potentially using significantly different terminology and conceptual frameworks to ground and disseminate their work. Such a lack of shared awareness and shared terminology may introduce problems for future research. Even the most basic terms used within this new wave of VAM-HRI work, that might naturally be assumed to have obvious, commonly agreed upon meanings --- words like ``virtual reality,'' ``augmented reality,'' ``mixed reality,''  ``user interface,'' and ``visualization'' --- are rendered imprecise by the multifarious uses of mixed reality visualizations and of virtual and mixed reality systems. It is thus critical for the research community to have a common set of terms and concepts that can be used to accurately and precisely describe and organize the wide array of work being done within this field. 

In this paper, we present a taxonomy for VAM-HRI 3D command sequencing paradigms and \emph{Virtual Design Elements (VDEs)} to provide researchers with a shared, descriptive basis for characterizing the types of systems HRI researchers are developing and deploying in both mixed and virtual reality systems. This taxonomy is the result of surveying 175 papers published in 99 conferences and journals over a time span of 30 years and is informed by recent (but nascent) attempts~\cite{williams2019reality} to grapple with the breadth and complexity of this field through the series of \textit{Virtual, Augmented, and Mixed-Reality for Human Robot-Interaction} (VAM-HRI) workshops held in conjunction with the ACM/IEEE International Conference on Human-Robot Interaction~\cite{williams2018virtual}. Each category, class, and VDE within our proposed taxonomy aims to provide HRI researchers (both those working within VAM-HRI and otherwise) with the shared language necessary to advance this subfield along a productive and coherent path.

Our goal in creating this taxonomy is not only to provide a shared language for researchers to use to describe and disseminate their work, but also to aid researchers in connecting with the host of complementary work being performed in parallel to their own. This taxonomy may enable researchers to better understand the benefits of different types of virtual imagery, more quickly identify promising graphical representations across different domains and contexts of use, and build knowledge regarding what types of representations may currently be under- (or over-) explored. 

As a final contribution of this work, we produce an online version of this taxonomy (\href{http://ibm.biz/vam-hri}{ibm.biz/vam-hri}) that visualizes the categorization of all 175 papers. Notably, this online platform enables any VAM-HRI researcher to issue a pull request to add their research to this categorization scheme. It is our hope that this platform serves as a living resource that the VAM-HRI community may use to track the continued progress and growth of our nascent field


\section{VAM-HRI Advancement Over Time}

To begin to understand how the field of VAM-HRI reached its current state, one can trace the development of VAM technologies back to Sutherland’s vision of “The Ultimate Display” (itself influenced by Vannevar Bush’s conception of the Memex) \cite{sutherland1965ultimate} and later developments with the Sword of Damocles system \cite{sutherland1968head}. The earliest major work leveraging VAM for HRI appears to date back to a push in the late 1980’s and early 1990’s with various work exploring robot teleoperation systems \cite{bejczy1990phantom, kim1987visual}. Perhaps the most fully-developed instance of these early systems was the ARGOS interface for augmented reality robot teleoperation \cite{milgram1993applications}. While the ARGOS interface used a stereo monitor, rather than the head-mounted displays in vogue today, the system introduced several design elements for displaying graphical information to improve human-robot communication and introduced concepts such as virtual pointers, tape measures, tethers, landmarks, and object overlays that would influence many subsequent designs. Later developments throughout the 1990’s introduced several other important concepts, such as the use of virtual reality for both actual robot control and teleoperator training \cite{hine1995vevi}, the integration of HMDs (including the first use of an HMD to control an aerial robot) \cite{de1997steering}, projective virtual reality where user “reaches through” a VR system to control a robot that manipulates objects in the real world \cite{freund1999projective}, the rise of VAM applications for robotics in medicine and surgery \cite{burdea1996virtual}, and continued work on ARGOS and ARGOS-like systems \cite{milgram1995telerobotic}. At a high level, major themes appear that focus on using VR for simulation or training purposes, VR and/or AR as new forms of information displays (e.g., for data from robot sensors), and VAM-based robotic control interfaces. While many of these developments appear initially promising, it is interesting to note that following an initial period of intense early research on HRI and VAM, later growth throughout the 1990’s appears to have happened at a relatively stable rate, rather than rapidly expanding. In addition, efforts to take research developments beyond laboratory environments into commercial/industrial systems appear to have been largely unsuccessful (indeed, even today robot teleoperation interfaces are still typically based on standard 2D displays rather than leveraging VAM).

In recent years, research in the field of VAM-HRI has seen explosive growth. This recent explosion is due in part to the emergence of commercial head-mounted displays (HoloLens, Vive, Oculus, etc.) as well as enhanced computer performance, which together have shifted the field from one requiring specialized hardware developed in research labs to one in which significant advances can be made immediately using standardized, inexpensive hardware systems that provide expansive software development libraries. 

The rise of new VAM-HRI research has demonstrated the potential for virtual reality (VR), augmented reality (AR), and mixed reality (MR) across a number of application domains that are of significant historical interest to the field of HRI, including education \cite{garcia2011mixed, chang2010improving}, social support \cite{quintero2015vibi, hedayati2018improving}, and task-based collaboration \cite{chandan2019negotiation, frank2016realizing}, while illustrating how MR and VR can be used as new tools in a HRI researcher's toolkit for robot design~\cite{cao2018ani}, human-subject experimentation~\cite{wijnen2020towards,williams2020using}, and robot programming and debugging~\cite{hoppenstedt2019debugging}.

AR visualization techniques offer significant promise in HRI due to their capability to effortlessly communicate robot intent and behavior. For example, researchers have shown how AR can be used to visualize various aspects of robot navigational state, including heading, destination, and intended trajectory \cite{walker2018communicating}. Similarly, others have shown how AR can be used to visualize a robot's beliefs and perceptions by displaying both exteroceptive (e.g., laser range finder data) and proprioceptive (e.g., battery status) sensor data~\cite{avalle2019augmented,cao2018ani}. Such virtual imagery can improve the situational awareness of human teammates during human-robot collaboration and teleoperation, making interaction more fluid, intuitive, safe, and enjoyable.

VR techniques, on the other hand, provide unique opportunities for safe, flexible, and novel environments in which to explore HRI. For instance, researchers have explored how virtual environments can help humans learn how to work with new robots~\cite{perez2019industrial} and help robots learn new skills~\cite{dyrstad2018teaching, iuzzolino2018virtual}. Using VR interfaces for these purposes reduces risks, addresses spatial and monetary limitations, and promotes new opportunities to experiment with new (or not yet physically feasible) robots and environments. Moreover, VR interfaces can be used to visualize the real world in new ways, allowing teleoperators and supervisors to view robots' real environments through video streams or point cloud sensor displays in a more immersive and intuitive manner than traditional 2D displays \cite{bosch2016towards, sun2020new}. In addition, VR systems may enable users to exert more fine-grained control over teleoperated robots, including unmanned aerial and nautical vehicles, by leveraging related technologies such as head tracking, haptic controllers, and tactile gloves~\cite{ibrahimov2019dronepick}. 


\section{VAM Interfaces for Robotics}

The advancements in hardware accessibility have created a host of new opportunities for exploring VAM technology as an interaction medium for enhancing various aspects of HRI. Various VAM displays, degrees of reality, and coordinate system calibration techniques, and VAM-HRI interface paradigms have all been used in different combinations to successfully apply VAM interfaces to robotics.


\subsection{The Reality-Virtuality Continuum}

VAM displays are computer displays that either immerse users in an entirely synthetic world or merge both the real world and a synthetic world. Note, that although VAM refers to more than just visual senses (i.e., haptic, auditory, etc.), for the duration of this paper VAM will refer only to synthetic imagery. 

VAM technology is capable of mixing varying degrees of reality and virtuality. As observed by Milgram et al. in 1994, all VAM displays fall upon a “Reality-Virtuality Continuum” \cite{milgram1994taxonomy}. This taxonomy has served as a useful tool for classifying VAM interfaces as well as coined the term “Mixed Reality.” Per the Reality-Virtuality Continuum, interfaces that place users in environments consisting of only synthetic imagery are considered “Virtual Reality” (VR), accordingly interfaces that only consist of real imagery are considered based in reality. This leaves a middle ground, where synthetic and real imagery are combined to form the space of “Mixed Reality” (MR). Within the design space of MR lie two sub-categories of “Augmented Reality” (AR), where synthetic imagery is added to a real environment, and “Augmented Virtuality” (AV), where real imagery is added to a synthetic environment. Together with VR, MR, consisting of AR and AV, one can categorize VAM interfaces and provide more specific design guidelines for HCI interfaces.

\begin{figure}
  \centering
  \includegraphics[width=\textwidth]{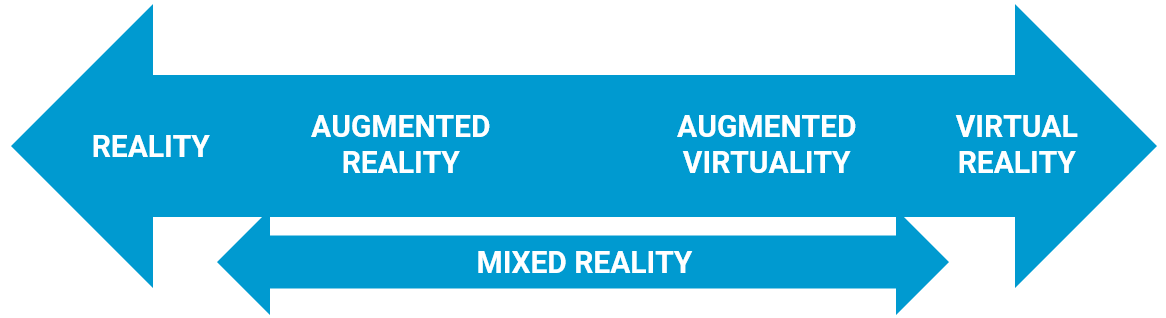}
  \caption{Milgram's Reality-Virtuality Continuum \cite{milgram1994taxonomy}.}
  \label{fig:continuum}
\end{figure}


\subsection{Display Hardware}

VAM interfaces can be implemented by various display hardware. The following is a list of common VAM display types.

\vspace{3mm}\noindent\textbf{2D Monitor Video Displays}: 2D monitor video displays provide a means of overlaying AR synthetic imagery on real images or video feeds, as ``window-on-the-world'' displays. Due to the inherent 2D nature of traditional monitors, users of these displays are not able to experience depth when viewing interfaces on these displays. Additionally, these displays tether users to computer terminals, and do not allow for a hands-free, free-roam of environments in which AR imagery are being added.

\vspace{3mm}\noindent\textbf{3D Monitor Video Displays}: 3D monitor video displays are similar to the above AR 2D monitor video displays, with the difference being that users can see interfaces with depth. Users can see with depth with either glasses-free 3D monitors or with 3D glasses (i.e., anaglyph, polorized, or active shutter).

\vspace{3mm}\noindent\textbf{Tablet Displays}: Tablet displays are also similar to AR 2D monitor video displays, although a major difference being users can freely explore their environments due to the portable nature of tablets. Additionally, tablets provide users with touch-based interactions, which is uncommon in the monitor-based displays.

\vspace{3mm}\noindent\textbf{Projectors}: Projector-based displays provide AR synthetic imagery to a user's environment, that users can see without the need of special equipment, such as glasses or head-mounted displays (HMDs), and also allow users to freely roam their environment. Drawbacks to this display type include: the projected imagery is inherently 2D; occlusions (from the environment, users, or robots) disrupt and/or block the AR imagery; and the projected imagery can get washed out in environments that are too bright.

\vspace{3mm}\noindent\textbf{CAVEs}: CAVEs, or Cave Automatic Virtual Environments, are VR displays where users are placed in a three to six walled environment. The walls of the environment display 3D imagery when paired with 3D glasses allowing users to be immersed in AV or VR environments. CAVE's allow for multiple users to experience the VR or AV environments, but unfortunately are expensive, confining, and immovable.

\vspace{3mm}\noindent\textbf{VR HMDs}: VR HMDs are 3D stereoscopic displays worn on users' heads to fully immerse users in VR or AV environments. Recent advances in VR HMD technology have allowed users to not only control the display with head motion, but with hand and body motion as well, permitting users to be hands-free while able to free to walk around synthetic environments naturally with their own body. A downside to these displays is that they enclose users off from the real world without an option of seeing reality from the display while worn.

\vspace{3mm}\noindent\textbf{Optical See-Through HMDs}: Optical See-Through HMDs present AR through transparent lenses, allowing users to see virtual imagery overlaid on the real world. Similarly, to VR HMD's, these displays provide users with hands-free, environmental free-roam experience. However, current technology is restricted to narrow field-of-views to see the virtual imagery. Additionally, the imagery provided by these HMDs are easily washed out in bright lights.

\vspace{3mm}\noindent\textbf{Video Pass-Through HMDs}: Video pass-through HMDs are unique in that they can provide AR, AV, and VR experiences to users. By mounting a stereo cameras to the front of a VR HMD, video pass-through HMD's can pipe video imagery from the real outside world to the dual lens within the HMD. The video feed can be intercepted between camera and lens, allowing for virtual imagery being added to the captured image frames. This process of displaying AR imagery allows these HMD displays to show AR imagery in environments with bright light, such as the outdoors.


\subsection{VAM Interface Coordinate Frame Rectification}

Finally, to successfully merge both reality and virtual reality in a single interface and to have VAM imagery appear in its appropriate position and orientation within the world, a singular coordinate frame must be maintained to manage the positions of the various virtual imagery. However, VAM interfaces for HRI present the unique challenge of multiple physical agents (users and robots) each having their own unique perspectives and coordinate frames. These agents' frames are also often moving and changing at any given times, requiring real-time tracking of users, robots, and objects within an environment. Research has explored the following various methods for extracting and unifying coordinate frames:

\vspace{3mm}\noindent\textbf{Fiducial Markers}: Fiducial markers are images that have their optical properties known by a VAM interface beforehand and act as visual reference points for the systems. When a VAM interface's camera detects a marker, it can determine the marker's relative pose to that of the camera. These markers are often placed in a robot's environment, on robots, or on objects within a robots environment. This method of frame rectification is popular due to its high portability and low-cost \cite{hashimoto2011touchme, borrero2012pilot, frank2017mobile, kobayashi2007overlay}.

\vspace{3mm}\noindent\textbf{Motion Capture Cameras}: Motion capture cameras can obtain poses of robots, users, and objects in an environment with high precision by tracking patterns of infrared reflecting markers. Unfortunately, these systems are expensive and immovable, often making interfaces that use this method of frame rectification constrained to laboratory environments \cite{walker2018communicating, hedayati2018improving, walker2019robot}.

\vspace{3mm}\noindent\textbf{Odometry}: Another method of unifying coordinate frames between user and robot is by using odometry (including visual odometry with computer vision algorithms such as SLAM \cite{mur2015orb}). If agent coordinate frames are initially synced at the initialization of a VAM-HRI interface, odometry can track the relative pose changes agents have undergone over time, which can be used to maintain a rectified coordinate system \cite{reardon2018come, gregory2019enabling}.

\vspace{3mm}\noindent\textbf{ICP Alignment}: Coordinate frame rectification is also made possible by using point cloud alignment algorithms, such as Iterative Closest Point (ICP) \cite{segal2009generalized}, which can take in separately collected point clouds and output the transformation between the two perspectives. This method requires all agents to have some overlap between their collected point clouds which makes this method most suitable for initial frame synchronization between multiple agents that is followed by odometry-based frame rectification methods above \cite{gregory2019enabling, reardon2018come}.

\vspace{3mm}\noindent\textbf{Machine Learning Image-Based Pose Estimation}: Relative poses between agents can also be found through machine learning algorithms, trained to estimate an object's pose through image frames. Current technology limits this method to known, simple objects; however, as machine learning algorithms strengthen \cite{xiang2017posecnn} it is feasible to imagine this method becoming more popular for agent frame rectification, especially if it means fiducial markers or motion capture cameras can be eliminated \cite{bolano2019transparent}.


\subsection{VAM-HRI Interaction Design Paradigms}
VAM-HRI interfaces often end up following higher-level implementation paradigms for various types of interactions. This can be seen especially in the cases of HMD teleoperation interfaces for remote robots and robot command sequencing.

\subsection*{HMD-Mediated Remote Robot Teleoperation Interfaces}
The recent advent of mass-produced HMD technology has seen a rise in HMD interfaces that mediate remote robot teleoperation, for either navigational tasks or manipulation tasks. As noted by Lipton et al. \cite{lipton2017baxter}, these interfaces can fall into three classes: Direct Interfaces, Virtual Control Room Interfaces, and Cyber-Physical Interfaces.

\vspace{3mm}\noindent\textbf{Direct HMD Teleoperation Interfaces}: Direct HMD teleoperation interfaces live-stream stereo video feeds from remote robots to the HMD's lenses. This process allows users to see from the robot's `eyes' with immersive 3D stereoscopic vision as if they were embodying the remote robot, especially if user head motions control robot head motions with a one-to-one mapping. These HMD-based interfaces have shown to significantly improve robot teleoperation tasks. Additionally, if virtual imagery is overlaid on the video stream, the interfaces become a AR interface \cite{higuchi2013flying}. However, there is a significant drawback associated with Direct HMD Interfaces that stems from both miscues from the user's proprioceptive system and the communication delays inherently found with current network technology. When a remote robot moves and the local user does not move, the user's proprioceptive system receives conflicting cues (visual cues of movement vs. no body motion detected) causing nausea. The same happens in reverse when a local user turns their head and robot's head does not immediately turn to match the movement (due to mechanical limitations or communication delays), which creates conflicting proprioceptive cues (no visual cues of movement vs. body motion detected) that also cause nausea.

\vspace{3mm}\noindent\textbf{AV HMD Teleoperation Interfaces}: To mitigate the nauseating effects of the Direct HMD Interfaces, two AV HMD Teleoperation Interface paradigms have arisen from research in the VAM-HRI field: Virtual Control Room and Cyber-Physical Interfaces \cite{lipton2017baxter}. In both interface styles, the state of the user's eyes are decoupled from the robotic systems state to remove the conflicting proprioceptive system cues. By placing the user in a AV environment, the users' eyes are represented by virtual cameras in the virtual space that move freely with the users head and body movements. This decoupling method helps mitigate nausea caused by communications/hardware delays and/or imperfect mappings between user head motion and robot head motion.

\begin{description}
  \item \textbf{Virtual Control Room Model}: In the Virtual Control Room Model, the user is placed in a virtual room that serves as a supervisory command and control center of a remote robot. Within the control room, the user is able to interacts with displays and objects in the virtual space itself, and can view 3D stereo video streams projected on the rooms walls thus still allowing the user to still see from the robot's perspective with depth \cite{lipton2017baxter, kot2014utilization} (see Figures \ref{fig:collection1}-A and \ref{fig:collection2}-C).
  \vspace{2mm}\item \textbf{Cyber-Physical Model}: In the Cyber-Physical Model a shared AV virtual space is created (typically with a one-to-one mapping) between: (1) a remote robot and a virtual environment; and (2) a human operator(s) and a virtual environment.
 Additionally, a 3D reconstruction of the robot's remote environment is rendered (typically with dense RGB point clouds) within the virtual environment to provide situational context and awareness to the human operator. A virtual robot replica of the remote physical robot is also added to the virtual environment in the same relative location within the virtual environment as in the real remote environment. This virtual robot mimics the remote real robot's state and actions. The user can also use the virtual robot to send commands to the remote real robot or visualize the current state or actions being undertaken by the physical robot. A benefit of this interface paradigm over that of the Virtual Control Room, is that the user can freely change their viewpoint within the remote environment by walking around the AV environment 3D reconstruction, as they are not only provided the view from the robot's camera(s). However, the sense of immersion from virtually embodying the remote robot is lost \cite{rosen2020communicating, sun2020new, allspaw2018remotely} (see Figure \ref{fig:collection1}-B).

\end{description}

\subsection*{VAM-HRI 3D Command Sequencing}
Our VAM-HRI literature review revealed recurring high-level themes for robot 3D command sequencing. We propose the following three paradigms that capture these methods of controlling robots with VAM interfaces: \emph{Direct Manipulation}, \emph{Environment Markup}, \emph{and Digital Twins}.

\vspace{3mm}\noindent\textbf{Direct Manipulation}: The paradigm of Direct Manipulation leverages more traditional methods to directly control robots and utilizes 3D translation and/or rotation input from either physical or virtual source, commonly to send teleoperation commands to end effectors or navigational systems. Direct manipulation from physical inputs include body/head tracking and VAM-based 3D controllers \cite{whitney2018ros}. Virtual controllers can act as metaphors for existing physical control input devices (i.e., levers, handles, joysticks, etc.) \cite{hashimoto2011touchme} or utilize novel designs unconstrained by physics such as floating control spheres \cite{krupke2016immersive}.

\vspace{3mm}\noindent\textbf{Environment Markup}: Under the Environment Markup paradigm, users send commands to robots by adding virtual annotations to a robot's environment. Examples of such environmentally-anchored annotations include waypoints, trajectories, and planned future poses of manipulable objects.  These annotations can take the form of a simple single command or can be chained/combined together to form a series of commands or a singular complex command \cite{chan2018multimodal, ishii2009designing}.

\vspace{3mm}\noindent\textbf{Digital Twins}: In contrast to commands sent under the Environment Markup paradigm, Digital Twin command sequencing does not add virtual annotations to a robot's environment. Instead, command sequences are generated by the manipulation of a \textit{digital twin} which is a virtual replica (or representation) of a real object or robot. For example, in the case of a robot with a digital twin, a user could press a virtual button on the robot's digital twin, at which point, the real robot would respond as if its own physical button was pressed. Additionally, robot teleoperation can be achieved by directly manipulating the robot digital twin (arm, body, etc.), after which, the real robot imitates the action taken by the digital twin (e.g., the robot moves itself in the exact trajectory taken by the digital twin or moves its end effector to the final position taken by the digital twin). Real object manipulation is performed in a similar manner, but in this case, a robot mimics a user's object digital twin manipulations on the associated real object \cite{sun2020new, frank2017mobile, hashimoto2011touchme, krupke2018comparison}.


\begin{figure}
  \centering
  \includegraphics[width=1\textwidth]{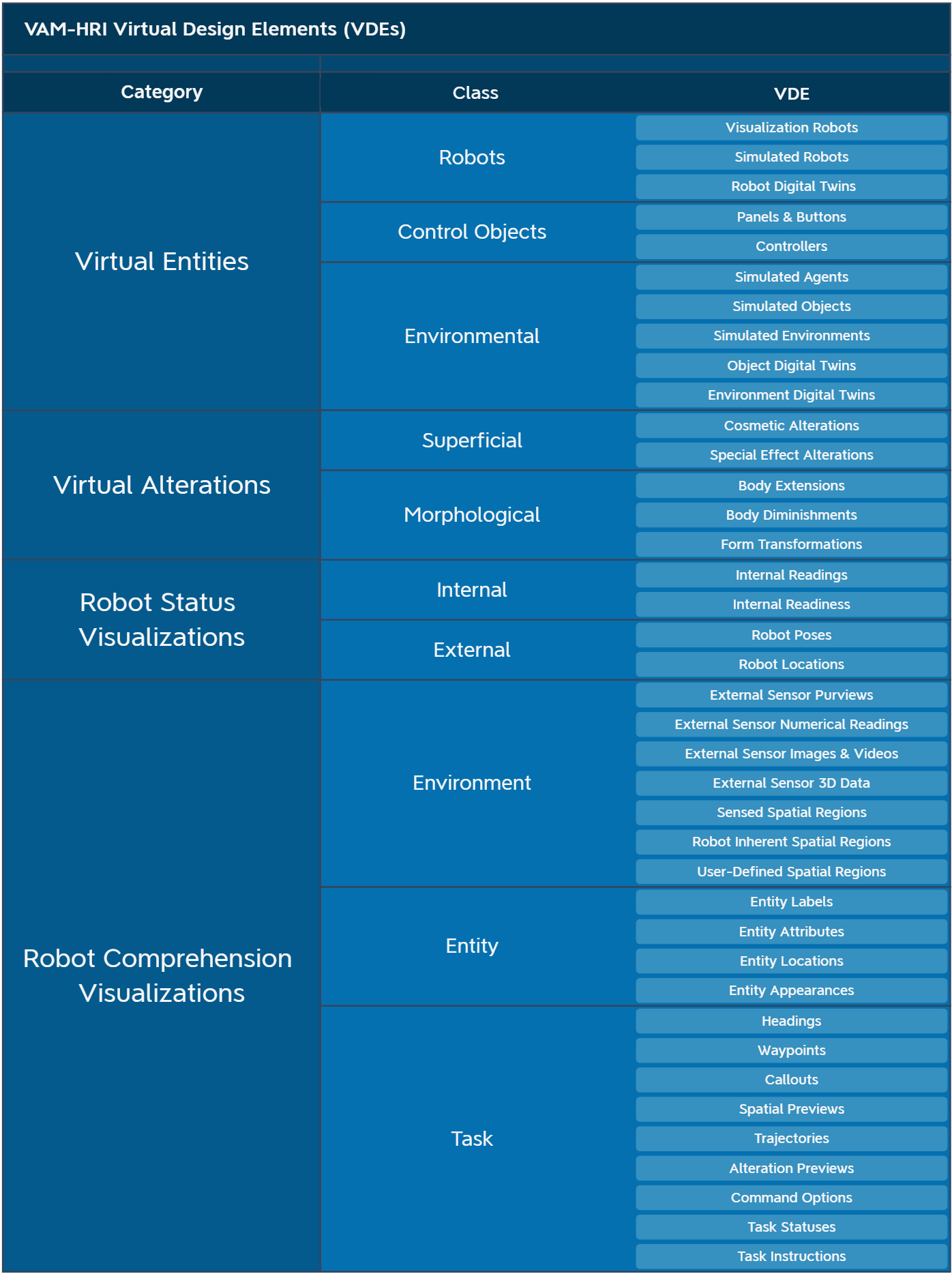}
  \caption{Virtual Design Element Taxonomy Table}
  \label{fig:Taxonomy}
\end{figure}

\section{VAM-HRI Virtual Design Element Taxonomy}

In this paper we propose a novel taxonomy for identifying and categorizing VAM-HRI Virtual Design Elements (VDEs)~\cite{williams2019reality}. VAM-HRI VDEs are VAM-based visualizations that impact robot \textit{interactivity} by providing new or alternate means of interacting with robots. VDEs can appear in two ways: (1) \emph{user-anchored} --- attached to points in the user's camera's coordinate system, unchanging as the user changes their field of view (see Figure \ref{fig:collection2}-D); or (2) \emph{environment-anchored} --- attached to points in the coordinate system of a robot or some other element of the environment, rather than the interface itself~\cite{williams2019reality} (see Figure \ref{fig:collection1}-C).

Over the course of the following sections, we will detail each category and class of VDE within the taxonomy, which we organize into the following categories:  \textit{Virtual Entities}, \textit{Virtual Alterations}, \textit{Robot Status Visualizations}, and \textit{Robot Comprehension Visualizations} (see Figure \ref{fig:Taxonomy}). Each VDE category was identified after surveying the aforementioned 175 VAM-HRI research papers, and acts as a high-level categorical grouping of VDEs that share common purposes for enhancing and/or manipulating human-robot interactions with virtual or mixed reality. Note that VDEs are \textit{instantiations} of the taxonomic sub-classes and can be used both in isolation and in synergistic conjunction by combining two or more VDEs (e.g., using a \emph{Cosmetic Alteration} to communicate the \emph{Internal Readiness} of a robot joint \cite{avalle2019augmented}).

\subsection{Virtual Entities}
The first category of VDEs we will examine are \textit{Virtual Entities}: visualizations in which virtual entities, such as robots or objects, are added to real or virtual user environments.  These VDEs can be a standalone VAM-based graphical elements that act as a visualization aid, input device, or simulation of an entity found in reality; however, they can also take the form of digital twins that are directly associated with a physical entity in the user and/or robot environment. We detail each of the three types of Virtual Entity classes below.

\subsection*{Virtual Entities -- Robots}
The \emph{Robots} class of the \emph{Virtual Entity} VDE category encompasses visualizations of robots that can be provided to users as either a visual tool for inspection or have full kinematic models and allow for complex interactions with users. These VDEs provide a level of immersion (a requirement for HRI simulations) unparalleled by simulations viewed on traditional 2D displays (such as Gazebo~\cite{meyer2012comprehensive}) since the Functional Virtual Robots is visually integrated within the user's environment. We identify three sub-classes of such virtual robot entities:

\vspace{2mm}\noindent\textbf{Visualization Robots} improve user understandings of a robot's current state or future actions but do not afford any two-way interactions with users (i.e., users cannot direct input to the graphical representation). These VDEs typically model a robot's 3D morphology, in whole or in part. Uses include providing users with a means of understanding a robot's current state in which case a virtual 3D model of a real robot mimics a physical robot's joint configurations in real time. Visualization Robots are particularly useful in situations of limited situational awareness when the user cannot directly see the physical robot (or portions of the robot), such as in remote teleoperation tasks \cite{kot2014utilization} (see Figure  \ref{fig:collection1}-A). Visualization Robot VDEs can also provide users with a preview of proposed or planned robot motion by overlaying a 3D robot model onto the environment to allow for better means to assess how a robot will navigate through an environment and whether it will successfully travel to a desired location without collisions. Finally, Visualization Robots can also show the locations of robots that are partially or fully occluded by the environment such as behind a wall or door \cite{rosen2020communicating} (see Figures \ref{fig:collection1}-B and \ref{fig:collection3}-D).

\vspace{2mm}\noindent\textbf{Simulated Robots} are not linked to a specific instance of a physical robot and are instead independent robot simulations used to evaluate robots in fully virtual settings when using a real robot is not ideal, such as when robot hardware is unavailable, unsafe, has limited battery life, and/or faces physical depreciation when operated repeatedly (as required for sample-inefficient learning algorithms and/or interaction studies). These VDEs are also used for evaluating simulated interactions with autonomous or manually teleoperated robots, e.g., in situations where collocated interaction is infeasible or hazardous for humans (i.e., working near large industrial robots, testing space exploration robots in zero gravity environments, etc.). Simulated Robots are also useful for user training without putting robot hardware at risk of being damaged by inexperienced users and for training real robots through virtual-to-real-world transfer learning techniques \cite{iuzzolino2018virtual} that require direct user interaction, such as learning from demonstration \cite{argall2009survey,de2018augmented, stilman2005augmented, meyer2018improving} (see Figure \ref{fig:collection1}-D).

\vspace{2mm}\noindent\textbf{Robot Digital Twins} operate in tandem with real robots and provide users with an immersive virtual robot that can be interacted with in lieu of a real, physical robot. By interacting first with a Robot Digital Twin, users can better predict how their actions will affect the system and offer foresight into the eventual pose and position of the physical robot as it mimics the actions taken by the virtual robot. For instance, mappings between the Robot Digital Twin and real robot include instantaneous duplication (the physical robot moves to match the Robot Digital Twin's position/attitude immediately), delayed duplication (the physical robot moves to match the Robot Digital Twin's position/attitude after a set period of time), confirmed duplication (the physical robot matches the Robot Digital Twin when triggered by the user), and more planning-oriented systems, such as using the Robot Digital Twin to denote waypoints or actions for future execution by the physical robot \cite{krupke2018comparison, sun2020new, walker2019robot, hashimoto2011touchme} (see Figure \ref{fig:collection1}-C).

\subsection*{Virtual Entities -- Control Objects}
These VDEs represent virtual objects that users can interact with to send direct commands that control robotic systems. Control Objects can be 2D or 3D and can be user-anchored (such as 2D buttons on an AR tablet that remain in static positions regardless of where the display is pointed --- see Figure \ref{fig:collection2}-D) or environment-anchored (such as a virtual 3D handles that remain fixed to a robot chassis --- see Figure \ref{fig:collection3}-F). We identify two sub-classes of Control Objects that allow for developers to rapidly prototype and evaluate various interfaces and designs for robot input without procuring or engineering actual hardware:

\vspace{2mm}\noindent\textbf{Panels \& Buttons} are virtual control objects that look and act like panels and buttons found in real life and 2D GUIs (e.g., buttons, sliders, switches, etc.) \cite{Li2019StarHopperAT} (see Figure \ref{fig:collection2}-D).

\vspace{2mm}\noindent\textbf{Controllers} emulate physical 3D input devices that leverages the 3D capabilities of VAM displays. Controllers often act as metaphors for existing physical control input devices (i.e., levers, handles, joysticks, etc.). However, Controller VDEs have also opened a nascent design space that allows robot designers to create interface input devices that are unconstrained by physics in the form of objects unable to be implemented or on earth or in reality, such as manipulable control toruses \cite{hashimoto2011touchme} or floating control spheres \cite{krupke2016immersive}, allowing for robot interactions that would otherwise be impossible to implement and/or evaluate with traditional non-VAM interfaces (see Figure~\ref{fig:collection3}-F).

\subsection*{Virtual Entities -- Environmental}
The class of Environmental \emph{Virtual Entities} encapsulate VAM-based visualizations of entities found in a user or robot's environment. For instance, virtual representations may be used to simulate agents, objects, and entire environments that do not physically exist in a robot's current development environment but will physically exist when a robot is deployed (e.g., to enable testing in a laboratory environment that mimics conditions a robot would encounter in the field as robots are unable to distinguish between the simulated virtual objects and real objects). In addition, these VDEs can be used for simulating how a robot might interact with such objects. These simulated entities often enhance interactions between robot and developers when debugging/assessing robotic systems by freeing developers from real agents/terrain to test a system. Alternatively, Environmental \emph{Virtual Entities} can be associated with physical agents, terrain, and objects already present in a robot's environment in the form of digital twins. We detail five VDEs in the Environmental sub-class of \emph{Virtual Entities}:

\vspace{2mm}\noindent\textbf{Simulated Agents} simulate, with virtual imagery, physical entities that normally have independent agency (e.g., autonomous robots, humans, animals, etc.). In the case of these simulations, the robot is not able to differentiate between real agents and Simulated Agents. The primary use of these VDEs is to enable robot testing and debugging without requiring the presence of key agents with whom robots would need to interact. For example, a large industrial robot might practice object handovers with a simulated human, without putting any human lives in harm's way, while the robot's developers observe and evaluate the mock interactions. Alternatively, a Simulated Agent VDE may be used in tandem with an autonomous Simulated Robot VDE for real humans to interact with, allowing for the testing of autonomous interactions with intelligent robots in situations where a physical robot is unavailable or currently infeasible~\cite{meyer2018improving} (see Figure \ref{fig:collection1}-D).

\vspace{2mm}\noindent\textbf{Simulated Objects} use virtual imagery to simulate the presence of physical objects in a robot's environment. It is important to note that these nonexistent virtual objects hold no association with any real objects in a robot's setting. These nonexistent objects can be used to simulate obstacles in debugging sessions with real robots (e.g., a virtual wall, table, chair, etc), without robot developers needing to procure physical objects, enabling rapid modification, deletion, and duplication of objects~\cite{borrero2012pilot} (see Figure \ref{fig:collection3}-G).

\vspace{2mm}\noindent\textbf{Simulated Environments} use virtual imagery to synthetically create the presence of environment areas or terrains. These VDEs can be used to evaluate autonomous robot responses to hazardous terrain (e.g., loose gravel, sand, water features, etc.) without endangering robots. Simulated Environments can also be used to evaluate robot interactions in environments that are difficult to find or recreate on Earth such as a lunar Moon with decreased gravity. Finally, Simulated Environments can provide a realistic setting to evaluate interactions (e.g., object hand-offs between user and robot) between Simulated Robots and users \cite{meyer2018improving}  (see Figure \ref{fig:collection1}-D).

\vspace{2mm}\noindent\textbf{Object Digital Twins} act as virtual replicas of associated real objects to sequence actions to be taken on their real-world equivalents. These VDEs can allow users to preview actions to be taken on the real object prior to robot execution. For example, a real cup to be moved by a real robot might have a virtual cup overlaid on its current position. A user could then interact with the virtual cup and move it to a new location, fine tune its final placement, and then command a robot to move the real cup to the position of the virtual cup. This interaction pattern can also enable robot action previewing similar to (and potentially in conjunction with) Robot Digital Twins \cite{krupke2018comparison, frank2017mobile} (see Figure \ref{fig:collection3}-H).

\vspace{2mm}\noindent\textbf{Environment Digital Twins} are virtual replicas of real environments rendered as a VAM-based visualization. These environments can be man-made structures/areas or outdoor terrain that are made to be exact replicas of their associated real world environment. As in Object Digital Twins, users could alter the state of the Environment Digital Twin, to have a robot take action on the real environment so its state matches its digital twin. Examples of such systems include interfaces that visualize real satellite terrain data as an Environment Digital Twin VDE to test and/or supervise aerial robot systems scouting a wildland forest fires across the associated real expanse of wilderness~\cite{omidshafiei2015mar} (see Figure \ref{fig:collection2}-B).

\subsection{Virtual Alterations}
The second category of VDEs we will examine are \textit{Robot Virtual Alterations}: graphical elements that allow a robot's appearance to become a design variable that is fast, easy, and cheap to prototype and manipulate. This category of VDE enables exciting new opportunities for HRI researchers and designers, especially since modifications to robot morphology are traditionally prohibitive due to cost, time, and/or constraints stemming from task or environment. We divide this category into classes involving (1) superficial alterations to robot appearance and (2) morphological alterations that substantially adjust robot form and/or perceived capabilities.

\subsection*{Virtual Alterations -- Superficial}
Superficial \emph{Virtual Alterations} use virtual imagery to change the appearance of physical parts of the robot. This change in appearance does not occur by altering the robot's form or morphology (i.e., adding a virtual arm, making the head invisible, etc.) but instead by changing the appearance of robots' physical surfaces or the space adjacent to those surfaces. We identify two sub-classes of such elements:

\vspace{2mm}\noindent\textbf{Cosmetic Alterations} alter the color, pattern, or texture of the robot's physical surfaces. The manipulation of robot surfaces enable new interaction patterns. These VDEs are considered cosmetic with respect to the robot’s morphology and can be combined with additional VDEs to provide a function. For instance, changing the color of a robot arm in a manufacturing context might call attention to a malfunctioning actuator, indicate a hot surface temperature, or discourage touching. In an educational setting, superficial alternations might change the texture of a robot arm to look soft or furry to encourage interaction with children. Additionally, as robots increasingly expand to new consumer domains in the near future, designers could use Cosmetic Alterations to make robots more eye catching in public spaces or enable end-user customization of personal robots in private living spaces to match home decor or personal taste \cite{avalle2019augmented} (see Figure \ref{fig:collection1}-E).

\vspace{2mm}\noindent\textbf{Special Effect Alterations} add virtual imagery around robots' physical surfaces to change their appearance indirectly. These effects can take various forms such as a glow effect added around a robot's body, virtual streamers that render behind a robot's arm as it moves, virtual flames that spray out of a robot's end effectors, or virtual light sources that indirectly alter the reflective appearance of a robot's physical surface. Although we did not come across this VDE during our literature survey, VAM-HRI is still a growing field and we envision this VDE holding value for manipulating human-robot interactions (i.e., adding virtual sparkles to a robot to make it more engaging to children in educational settings).

\subsection*{Virtual Alterations -- Morphological}

\begin{figure}
  \centering
  \includegraphics[width=.77\textheight]{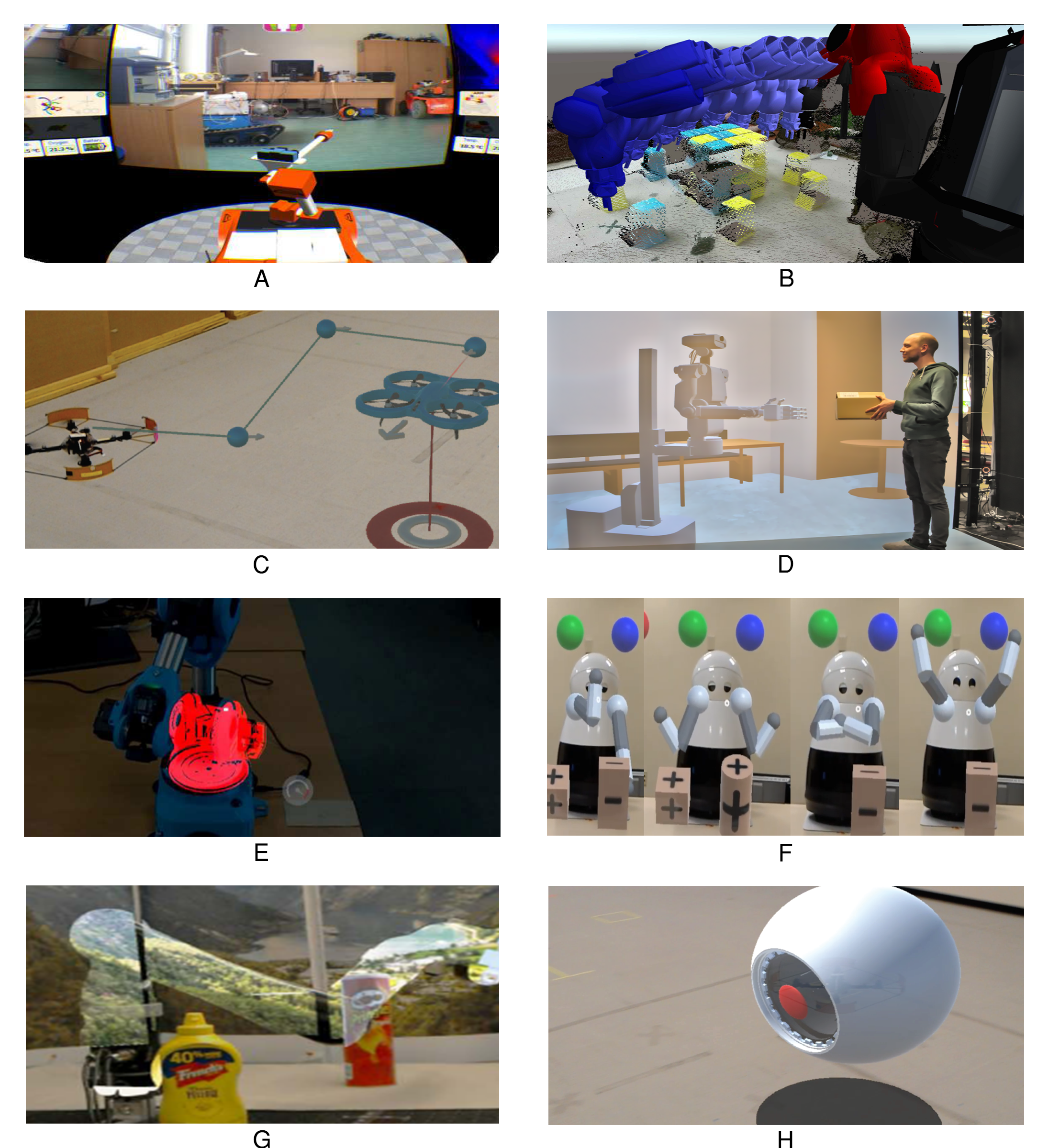}
  \caption{A: Virtual Control Room, Visualization Robot, and Internal Readings \cite{kot2014utilization}, B: Cyber-Physical Interface, Visualization Robot, Spatial Previews, External Sensor 3D Data, and Trajectories \cite{rosen2020communicating}, C: Robot Digital Twin, Headings, Waypoints, and Trajectories \cite{walker2019robot}, D: Simulated Robots, Simulated Agents, and Simulated Environment \cite{meyer2018improving}, E: Cosmetic Alterations and Internal Readiness \cite{avalle2019augmented}, F: Body Extensions and Callouts \cite{groechel2019using}, G: Body Diminishment \cite{taylordiminished}, H: Form Transformation and Heading \cite{walker2018communicating}.}
  \label{fig:collection1}
\end{figure}

Morphological \emph{Virtual Alterations} connect or overlay virtual imagery on a robot platform to fundamentally alter a robot's perceived form and/or function by creating new ``virtually/physically embodied'' cues, where cues that are traditionally generated using physical aspects of the robot are instead generated using indistinguishable virtual imagery. For example, rather than directly modifying a robot platform to include signaling lights as in \cite{szafir2015communicating}, an AR interface might overlay virtual signaling lights on the robot in an identical manner. Alternatively, virtual imagery might be used to give anthropomorphic or zoomorphic features to robots that don't have this physical capacity (e.g., adding a virtual body to a single manipulator or a virtual head to an aerial robot). Virtual imagery might also be used to obscure or make more salient various aspects of robot morphology based on user role (e.g., an override switch might be hidden for normal users but visible for a technician). These alterations may also enable new forms of interaction not previously possible for a given morphology, such as enabling functional robots to provide gestural cues \cite{szafir2015communicating}. We identify three sub-classes of such morphological \emph{Virtual Alterations}:

\vspace{2mm}\noindent\textbf{Body Extensions} add virtual parts to a robot without changing its underlying form such as an aerial robot is still recognizable as a UAV even if a virtual arm is added to its chassis, which would not be the case if virtual imagery were overlaid on the aerial robot to make it look like a floating robotic eye \cite{walker2018communicating} (see Figure \ref{fig:collection1}-H). Extensions do not necessarily need to be traditional robot parts and might instead appear as human heads, animal limbs, or even imagined parts like magical wings \cite{cao2018ani, groechel2019using} (see Figure \ref{fig:collection1}-F).

\vspace{2mm}\noindent\textbf{Body Diminishments} visually remove, rather than add, portions of a robot (e.g., grippers, heads, arms, wheels) through diminished reality (DR) techniques \cite{mori2017survey}. An important use of this VDE is to resolve teleoperation occlusions that occur when a robot arm blocks the line-of-sight between its camera and the object being manipulated~\cite{taylordiminished} (see Figure \ref{fig:collection1}-G).

\vspace{2mm}\noindent\textbf{Form Transformations} overlay virtual imagery onto real robots to change the robot's underlying form and/or make it appear as something other than a robot entirely. Similar to Robot Diminishments, this VDE can change the form of the robot to make it more or less appealing to a targeted user-groups or utilize new communication methods, depending on the designer's intentions. These form alterations need not be limited to that of new mechanical forms, but can include any form such as that of a human, animal, or fictional character, all of varying degrees of realism \cite{walker2018communicating, zhang2019evaluation} (see Figures \ref{fig:collection1}-H).

\subsection{Robot Status Visualizations}
The third category of VDEs we present are \textit{Robot Status Visualizations}: a set of elements focused on enabling designers to rapidly and easily assess the current state of a robot. We divide this category into classes focused on (1) internal and (2) external robot status.

\subsection*{Robot Status Visualizations -- Internal}
Internal \emph{Robot Status Visualization} VDEs convey internal sensor readings and/or the operational status of robot sensors and actuators. We identify two sub-classes of such elements:

\vspace{2mm}\noindent\textbf{Internal Reading} VDEs display data returned \emph{from} internal sensors (e.g., battery levels, robot temperatures, wheel speeds). Making this information more readily available to users may enhance situational awareness and prevent mishaps such as robots running out of battery in the middle of mission-critical tasks or traveling too fast through environmental hazards~\cite{kot2014utilization} (see Figure \ref{fig:collection1}-A).

\vspace{2mm}\noindent\textbf{Internal Readiness} VDEs display data \emph{about} sensors and actuators, such as whether a sensor is ready to collect data or whether an actuator is ready to function, and if not, why (e.g., whether a sensor is disconnected, an actuator is experiencing a fault, etc.). Such information may improve debugging and help prevent robots from operating as black-boxes, with users left wondering why a robot is not functioning as expected \cite{avalle2019augmented} (see Figure \ref{fig:collection1}-E).

\subsection*{Robot Status Visualizations -- External}
External \emph{Robot Status Visualization} VDEs communicate the robot's external state (as the robot perceives it) by providing information regarding its current pose and location. We identify two sub-classes of such elements:

\vspace{2mm}\noindent\textbf{Robot Pose} VDEs convey a robot's knowledge of its own pose (i.e., configuration and orientation). These VDEs can take different forms such as a textual display of numerical joint angles, a 3D model of a State Visualization Virtual Robot, or a rendering of a virtual axis anchored to a robot's joints~\cite{kot2014utilization, nawab2007joystick} (see Figure \ref{fig:collection2}-A).

\vspace{2mm}\noindent\textbf{Robot Location} VDEs convey where a robot is in the environment, e.g., as an occluded robot's outline with a Spatial Visualization Virtual Robot or a virtual indicator on the other side of a wall or door, as a top-down radar-like display showing user and robot locations, or as an off-screen indicator to direct a user's attention to a robot outside the current field-of-view~\cite{chandan2019negotiation, walker2018communicating} 
(see Figure \ref{fig:collection3}-D).

\subsection{Robot Comprehension Visualizations}
The fourth and final category of VDEs we examine are \textit{Robot Comprehension Visualizations}: visualizations that convey what a robot believes about its environment, and its current or planned task. VAM-HRI VDEs present a powerful medium for conveying this information, as VDEs can be directly overlaid on a robot's environment. For example, a visual trajectory spline rendered on a floor can wrap around a wall, indicating not only that the robot sees the wall, but also that the robot's planned actions will avoid the wall. 

\subsection*{Robot Comprehension Visualizations -- Environment}
Environment-based \emph{Robot Comprehension Visualization} VDEs communicate to the user what the robot believes about its environment, including where environment information is being collected from, what environment information has been collected, and/or what a robot has inferred from such information. We identify seven sub-classes of such elements:

\vspace{2mm}\noindent\textbf{External Sensor Purviews} are environment-anchored visualizations that show users where a robot's external sensors (LiDAR, cameras, etc.) are collecting data and/or how and where those sensors are positioned \cite{hedayati2018improving, kobayashi2007overlay} (see Figure \ref{fig:collection2}-B).

\vspace{2mm}\noindent\textbf{External Sensor Numerical Readings} convey numerical data returned from a robot's external sensors. These readings can be user-- or environment--anchored and can be shown explicitly with digits or as more abstract visualizations such as progress bars, virtual thermometers, or virtual weight scales \cite{lipton2017baxter, cao2018ani} (see Figure \ref{fig:collection2}-C). Note that the external sensors need not physically attached to a robot, such as data from motion capture cameras regarding the distance between a robot and a nearby object.

\vspace{2mm}\noindent\textbf{External Sensor Images \& Videos} allow the user to see remote environments or to see from a robot's perspective. Images and videos can be presented as either user- or environment-anchored visualizations from cameras on the robot or in the robot's environment. When stereo cameras are paired with a stereo interface (e.g., an HMD, 3D monitor, or CAVE), users can see the images and videos with depth, granting enhanced immersion and teleoperation capability~\cite{lipton2017baxter, kot2014utilization, Li2019StarHopperAT} (see Figure \ref{fig:collection2}-D). 

\vspace{2mm}\noindent\textbf{External Sensor 3D Data} convey depth information to recreate remote robots' environments. These reconstructions can take various forms (e.g., point clouds, voxel maps, 3D meshes, etc.) and aim to present sensed depth data in a manner that allows users to perceive remote environments as if they were there in-person. Unlike Images and Video VDEs, the 3D reconstructions utilized by External Sensor 3D Data VDEs enable immersive and free exploration of a remote robot's environment, without being restricted to the robot's perspective~\cite{sun2020new, rosen2020communicating} (see Figure \ref{fig:collection1}-B). As in the case of the previous two VDEs, the external sensors do not need to be physically attached to the robot and this VDE includes 3D data collected from user-worn HMDs, such as the spatial map generated from a HoloLens.

\vspace{2mm}\noindent\textbf{Sensed Spatial Regions}, the first of three region-based VDEs, visualize regions that a robot has identified within its its environment. These visualized regions are produced by a robot through the analysis of sensed environment data, such as exploration frontiers or traversable vs. non-traversable areas depicted through an occupancy grid (unlike 3D reconstructions that do not perform any logical analysis and categorization of environmental regions). Regions can be annotated with information (i.e., estimated information gain if the area were to be explored) \cite{reardon2019augmented} (see Figure \ref{fig:collection2}-E).

\vspace{2mm}\noindent\textbf{Robot Inherent Spatial Regions} are not informed by data sensed from the environment, but are instead inherent to the robot based on its form or operational mode. For example, these may depict the regions a robot can physically reach or show areas a robot may operate best in, such as the optimal area in which to perform an object handover \cite{frank2017mobile} (see Figure \ref{fig:collection2}-F).

\vspace{2mm}\noindent\textbf{User-Defined Spatial Regions} are regions defined by user input, rather than by the robot's sensors or physical constraints (e.g., a user drawing a bounding box on the floor of a robot's environment with a tablet display). These regions have many potential uses, but are most commonly used to define areas a robot should not enter, in the form of virtual boundaries \cite{sprute2019study} (see Figure \ref{fig:collection2}-G).

\subsection*{Robot Comprehension Visualizations -- Entity} 
\begin{figure}
  \centering
  \includegraphics[width=.77\textheight]{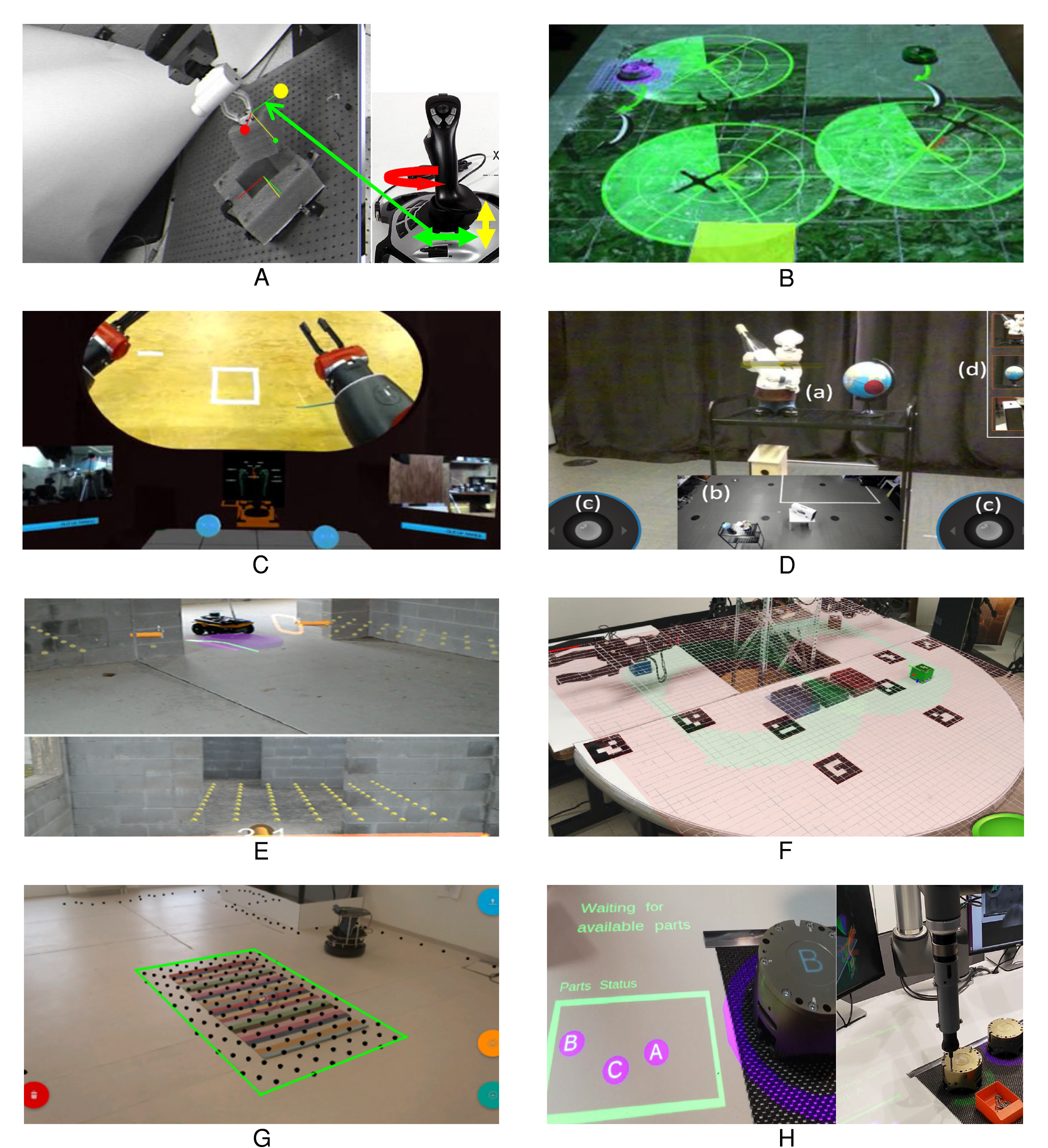}
  \caption{A: Robot Poses \cite{nawab2007joystick}, B: External Sensor Purviews and Environmental Digital Twin \cite{omidshafiei2015mar}, C: Virtual Control Room and External Sensor Numerical Readings \cite{lipton2017baxter}, D: External Sensor Images \& Videos and Virtual Panels \& Buttons \cite{Li2019StarHopperAT}, E: Sensed Spatial Regions \cite{reardon2019augmented}, F: Robot Inherent Spatial Regions \cite{frank2017mobile}, G: User-Defined Spatial Regions \cite{sprute2019study}, H: Entity Labels, Entity Locations, and Task Status \cite{bolano2019transparent}.}
  \label{fig:collection2}
\end{figure}

Entity-based \emph{Robot Comprehension Visualization} VDEs convey what a robot knows or believes about an entity (i.e., an object, human, another robot, etc.), such as where an entity is, what an entity is, and attributes an entity holds. We identify four sub-classes of such elements:

\vspace{2mm}\noindent\textbf{Entity Labels} act as identifiers for entities known by a robotic system. These visualizations enable users to easily reference entities in spoken commands without the need for referring expressions such as enabling users to instruct robots through commands such as ``pick up cube B'' \cite{bolano2019transparent, sibirtseva2018comparison} (see Figure \ref{fig:collection2}-H). Additionally, these VDEs allow robots to label points of importance in the environment such as a room's primary access point \cite{chandan2019negotiation} (see Figure \ref{fig:collection3}-D)

\vspace{2mm}\noindent\textbf{Entity Attributes} convey information a robot knows about an entity's characteristics, such as whether an entity is heavy, delicate, or dangerous; information known about the entity's affordances; the current state of an entity (e.g., an entity that's too hot, in a dangerous location, still drying, sleeping, charging its battery, etc.); or an entity's geometry and shape (e.g., optimal grasp points or surface normals)~\cite{chan2018multimodal} (see Figure \ref{fig:collection3}-A). 

\vspace{2mm}\noindent\textbf{Entity Locations} highlight the locations of entities within the robot's environment through rings, arrows, bounding boxes, etc. This VDE is especially useful when the location of an entity is occluded by walls or containers or outside of the user's field-of-view~\cite{dima2020joint}. These can also be used to highlight task- and dialogue-relevant entities, either by allowing robots to passively highlight entities that are of interest to the current task or the subject of the robot's current attention, or to actively call interlocutors' attention to entities in the same way that humans typically would through deictic gaze and deictic gesture~\cite{williams2019mixed, sibirtseva2018comparison, quintero2015vibi, bolano2019transparent} (see Figures \ref{fig:collection2}-H).

\vspace{2mm}\noindent\textbf{Entity Appearances} show what an entity looks like, unlike Entity Location VDEs that communicate the location of an entity. These VDEs are primarily used when an entity is occluded from view and draws analogies to ``X-ray vision'' by showing users the appearance of real-life entities that are partially or fully visually hidden within the users' environment (e.g., inside a box, behind a robot arm, or on the other side of a wall or door). This interface feature may be particularly useful in environments, such as warehouses, where objects are stored in sealed containers with contents only knowable through data representations that are exclusively computer-readable (e.g., barcodes) \cite{ganesan2018better} (see Figure \ref{fig:collection3}-B).

\subsection*{Robot Comprehension Visualizations -- Task}
Task-based \emph{Robot Comprehension Visualization} VDEs display what a robot understands about its current or planned task, including where to move, how to move, what objects to act upon, and how to act on those objects. These VDEs can also convey information regrading general task understanding, such as task status and outcomes. We identify nine sub-classes of such elements:

\vspace{2mm}\noindent\textbf{Headings}, the first type of VDE in this class, do not show the actual path the robot or its manipulators will take, but simply the direction they are currently traveling in or will be traveling next. These visualizations commonly take the form of arrows pointing in the direction of planned movement; however, they can also take more unique forms (i.e., the utilization of a Form Transformation VDE to provide an eyeless robot with virtual eyes that designers can use to provide gaze cues that communicate future movement intentions \cite{walker2018communicating} (see Figure \ref{fig:collection1}-H). Headings may be useful for autonomous robots in crowded, shared spaces with human pedestrians or dynamic obstacles, in which navigational plans need to be recalculated by the robot quickly and frequently. Researchers have also shown how headings can be particularly effective when displayed using projectors, enabling all bystanders to see the intended movements of robots without observers needing to each wear or use specialized hardware~\cite{shrestha2018communicating} (see Figure \ref{fig:collection1}-C).

\vspace{2mm}\noindent\textbf{Waypoints} are environmentally-anchored visualizations of intermediate navigation points. These are typically used to visualize robot intentions but can also be used by robots to suggest spatial destinations indicating where users should move. Waypoints provide another method of previewing robot motion and can either be automatically placed in an environment by a robot trajectory planner or manually placed in the environment by a user. Waypoints are also often combined with other VDEs to show additional information known about each waypoint or what will be performed at each waypoint~\cite{chan2018multimodal, walker2018communicating, walker2019robot} (see Figure \ref{fig:collection1}-C).

\vspace{2mm}\noindent\textbf{Callouts} are visualizations that communicate where a user should focus their attention. These VDEs use visualizations to attract attention to an object or location, such using a virtual arrow to show where a robot heard a sound or pointing at an object a user should look at \cite{groechel2019using} (see Figure \ref{fig:collection1}-F).

\begin{figure}
  \centering
  \includegraphics[width=.77\textheight]{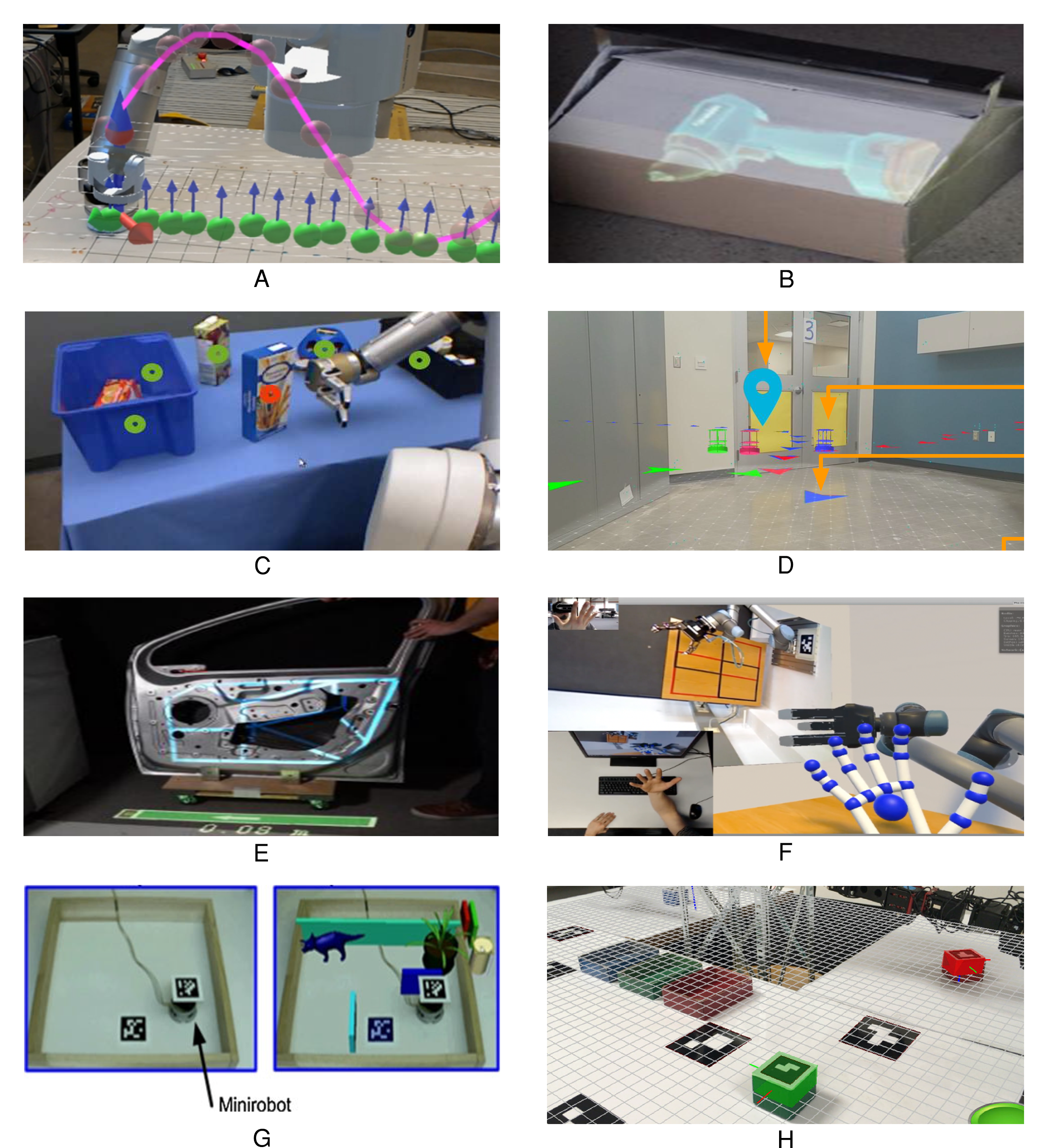}
  \caption{A: Entity Attributes and Alteration Previews \cite{chan2018multimodal}, B: Entity Appearances \cite{ganesan2018better}, C: Command Options \cite{quintero2015vibi}, D: Visualization Robots, Robot Locations, and Entity Labels \cite{chandan2019negotiation}, E: Task Instructions \cite{ganesan2018better}, F: Virtual Controllers \cite{krupke2016immersive}, G: Simulated Objects \cite{borrero2012pilot}, H: Spatial Previews and Object Digital Twins \cite{frank2017mobile}.}
  \label{fig:collection3}
\end{figure}

\vspace{2mm}\noindent\textbf{Spatial Previews} use environment-anchored visualizations to show future poses of robots, objects, and other environmental entities. These VDEs can explicitly communicate the expected future position and/or orientation of an entity during a task. A common use for these previews is to depict where a robot will move or where a robot will move an object during a manipulation task. However, robots can also use these VDEs to make requests to users by indicating where a user should place an object. These VDEs can be depicted in various ways, including 2D circles on the ground, complex 3D wireframe or shaded models, combining waypoints with flags indicating orientation at those waypoints, or with one or more Spatial Visualization Virtual Robots~\cite{rosen2020communicating, frank2017mobile} (see Figures \ref{fig:collection1}-B \& \ref{fig:collection3}-H). 

\vspace{2mm}\noindent\textbf{Trajectories} display spatial paths that a robot intends to follow or that it believes an object or agent will follow. These environment-anchored visualizations can show both robot navigation paths and manipulator paths. Trajectories can be visualized in various ways such as lines, splines, or dense stroboscopic Future Robot Pose VDEs in the form of State Visualization Virtual Robots. For example, one common implementation of Trajectory VDEs consists of rendering a trajectory for each wheel on a ground robot, which helps users to anticipate whether or not a wheel will collide with or fall into a terrain hazard. Trajectories can be used to not only indicate the path a robot will take in the future, but also the path a robot (or human) has taken in the past. Trajectory VDEs are typically only used to enhance a user's view into a robot's internal model but can also be used to provide new opportunities for control over the robot (e.g., a user directly manipulating the trajectory visualization by grabbing the trajectory line or spline and moving it with their hands). Finally, trajectories can encode data along their paths, such as robot velocities \cite{walker2018communicating, rosen2020communicating, leutert2013spatial, walker2019robot} (see Figure \ref{fig:collection1}-C).

\vspace{2mm}\noindent\textbf{Alteration Previews} show the intended permanent modifications a robot plans to make on an object. These rendered previews give the user a chance to verify if the robot's plan matches that of the user's task goal(s) prior to execution and cancel or modify them if needed. These modifications typically show how an object will appear during or after the modifications take place (e.g., displaying the proposed path of an etching tool on an object's surface, where holes will be drilled, how a wall will look after being painted, how a steel bar will look after being bent, etc.). Additionally, these visualizations may communicate what actions will be applied to an object, such as varying pressures applied along the surface on object. This type of VDE is especially useful in circumstances where robot errors arising from command misinterpretation mean the object being acted upon will no longer be usable, potentially wasting hours or days of time and resources to replace the incorrectly modified object~\cite{chan2018multimodal, leutert2013spatial} (see Figure \ref{fig:collection3}-A).

\vspace{2mm}\noindent\textbf{Command Options} present to users what actions a robot can (or cannot) take in a given state. This may take the form of a robot displaying virtual imagery that indicates what object(s) it can currently pick and/or potential grasp points the robot can utilize \cite{quintero2015vibi}. In addition to showing user what actions a robot can take, these VDEs also allow a robot to inform users that it is incapable of performing an action it believes the user wishes or might wish it to perform, sometimes with an explanation as to why they are not possible. These VDEs may reduce user frustrations with robotic systems by avoiding situations where a robot silently fails to execute commands and improve user efficiency by aiding users in preemptively realizing that a task will not be performed correctly (or at all)~\cite{arevalo2020there} (see Figure \ref{fig:collection3}-C).

\vspace{2mm}\noindent\textbf{Task Status} convey beliefs regarding the status of a task currently or previously executed. These VDEs may be represented as traditional textual or numerical visualizations, or as more abstract visual representations, such as progress bars. These visualizations can facilitate human-robot collaborative task planning by helping users quickly and easily understand the current task state a robot is executing, improve debugging by enabling users to compare the state a robot thinks it is in with its actual state, or how much longer a task will take to complete \cite{bolano2019transparent, walker2018communicating, ganesan2018better} (see Figure \ref{fig:collection2}-H). Additionally, these VDEs can also convey the status of a concluded task, whether it has resulted in success, failure, or error. These visualizations help users understand what a robot believes the outcome of a task to be (even if incorrect, which may aid in robot debugging). These visualizations can inform a user as to why a task resulted in failure or error, which can often be a mystery to users who would otherwise need to consult complex error logs~\cite{ganesan2018better, de2018augmented}.

\vspace{2mm}\noindent\textbf{Task Instructions} enable humans and robots to effectively instruct and guide each other by communicating next steps in collaborative tasks. These instructions can take various forms such as explicit instructions written in text or more abstract instructions that inform a user what to do to accomplish a task such as using an Robot Extension Morphological Alteration VDE to add virtual arms to a robot that point at an object for a user to interact with~\cite{groechel2019using, ganesan2018better} (see Figures \ref{fig:collection3}-E).

\section*{Robot Functionalities and Domains Supported by VAM-HRI}

Now that we have described the space of Virtual Design Elements used in VAM-HRI applications, we are ready to examine the ways in which these different VDEs have been used to enhance certain robot functionalities as well as the application domains in which those VAM-enhanced robotic functionalities have proven most useful.


VAM-VDEs have been leveraged to enhance a number of fundamental robotic needs. We divide these robotic functionalities into the following categories: navigation, object manipulation, prototyping, human training, robot training, debugging, swarm supervision, and social interaction. In this section we will describe the way in which VAM VDEs have been used for each of these purposes.

\subsubsection{Robot Navigation} 
For robotic applications that do not involve stationary robots, indoor and outdoor environments of varying scale must be safely and efficiently navigated. Our survey of the literature identified 51 papers in which AR and VR were used for this purpose. Kästner et al. \cite{kastner2019augmented}, for example, presents an approach in which \textit{trajectories} are used to visualize a robot's intended navigation path within human teammates' HMDs. Moreover, Kästner et al. allow users to re-specify the robot's destination through manipulation of a virtual arrow used to denote its destination. Similarly, Stotko et al. \cite{stotko2019vr} present a VR-based system for remote robotic teleoperation, in which robotic sensor data is visualized directly within the user's HMD to allow them to explore the robot's environment through that robot's perspective with high levels of immersion. 

\subsubsection{Robot Manipulation} 
While not necessary in some social domains, most task-oriented robotics applications require robots to physically manipulate objects in their environment, e.g. in assembly or sorting tasks. Our survey of the literature identified 105 papers in which AR or VR were used for this purpose.  Many of these papers used AR in order to help human teleoperate robotic manipulation more accurately and safely. Krupke et al.~\cite{krupke2018comparison}, for example, present a system that superimposes a virtual Robot Digital Twin over a real robot in augmented reality. Users of this system can then control the real robot to perform a pick and place task through manipulation of the Robot Digital Twin and Object Digital Twins shown in their HMD. After each user command, the virtual robot simulates performance of the commanded task. If the user is satisfied, they can then trigger the real robot to perform that command. While in this use case, the robot operator is present alongside the robot, VAM techniques can also be used to help users to remotely control robot manipulation. Naceri et al. \cite{naceri2019towards}, for example, present a VR interface for real-time robot teleoperation. In this interface, the remote environment is visualized using streaming External Sensor Images and Videos and External Sensor 3D Data, to make robot teleoperation more effective. 

\subsubsection{Robot Training}
In many circumstances, robot end-users are tasked with teaching and/or programming a more general-purpose robot to perform a specific job, such as how to sort dishes, open and pour a bottle, or build furniture. More traditional robot training methods, such as learning from demonstration \cite{argall2009survey}, can be especially challenging tasks when minimal feedback from the robot is provided to the user regarding how well the robot is learning, why the robot is not learning, or how to provide teaching inputs to the robot. VR and AR can be used in these contexts to visualize task-relevant objects and obstacles as virtual objects. Our survey of the literature identified 31 papers that used VR or AR for such purposes. Sprute et al.~\cite{sprute2019learning}, for example, present an AR system for teaching robots the extent of its operating environment though User-Defined Spatial Regions that serve as virtual walls. The define virtual borders are defined with an AR tablet interface, which the robot uses to subsequently generate its navigation plans.

\subsubsection{Human Training} 
Alternatively, in some domains, humans need to be trained to operate robots within a safe and constrained environment. Additionally, robots may need to share task-relevant information with their human teammates, such as instructions on how to complete a human-robot shared task.Our survey of the literature identified 35 papers in which AR and VR were used for this purpose. In these cases, VR and AR are often used to develop personalized, low-cost training environments. Many of these use cases are in the context of training users to use vehicles (autonomous or otherwise)~\cite{ropelato2018adaptive, arppe2019uninet} or in the context of training humans to train robots~\cite{gadre2019end}. Driving simulations provide learners a safer place to improve their skill without worrying about causing disturbances to others.

\subsubsection{Robot Debugging} 
Similarly, even outside of training procedures, robot programmers as well as end users often need to determine on the fly why a robot is acting in a particular way, especially when unexpected behavior is displayed. Current robots typically require these users to parse detailed error logs to answer fairly simple questions, such as why a robot's end effector stopped moving. This sort of question can have multiple answers, ranging from motor faults to payloads that exceed maximum weights.  Our survey of the literature identified five papers in which AR and VR were used to enhance robot debugging. VAM techniques can be added to robotic systems to aid in quickly answering common debugging questions. For example, robot debugging can be enhanced by rendering virtual imagery to localize and efficiently explain robot faults. VAM-HRI systems have been designed that visualize robotic faults within user HMDs through the use of Virtual Cosmetic Alterations on the robot that highlight the robot parts that are currently experiencing faults with paired visualizations that provide at-a-glance information about fault type (e.g., sensor fault, servo fault, end effector overloading, etc.) \cite{avalle2019augmented, de2018augmented}. 

It is important to note that HRI is not restricted to interactions between robots and end-users, but between robots and developers as well. When going through the process of creating a new robot or robot algorithm, robot designers go through extensive cycles of debugging. To iterate and improve a robotic system these users must understand why a robot/algorithm is not performing as expected. During this testing phase VAM-HRI techniques can allow robot designers to more easily see what a robot is thinking through virtual imagery, such as if a robot detects an obstacle while testing an autonomous navigation algorithms \cite{kobayashi2007overlay}.

\subsubsection{Robot Prototyping} 
Similarly, while many of the previous use cases have focused on on-line robot tasking, VAM can also be used in the initial design of robots before they are ready for deployment. When VAM is used for robot prototyping, virtual imagery is used to preview robot designs and/or functionality. This virtual imagery can be used either to represent a completely virtual robot or add virtual parts to a robot - all without using physical robotic hardware. VAM robot prototyping saves on both the monetary costs of robotic hardware as well as hours of labor that would otherwise be needed to install or program robot parts during the design process. Our survey of the literature identified 10 papers in which AR or VR were used for this purpose. Cao et al. \cite{cao2018ani}, for example, introduce a mixed reality robot prototyping system for people building DIY robots that allows users to virtually assemble and construct robots in AR with Visualization Robots and Body Extensions. Using this system, hobbyists can test their designs in mixed reality before executing those designs in real life with Simulated Robots and Simulated Objects.

\subsubsection{Social Interactions} 
In the domain of social robotics, robot developers use a variety of design strategies to manipulate users' perceptions of robots as being more trustworthy, engaging, and/or approachable. Our survey of the literature identified 17 papers in which AR or VR were used for this purpose. Many of these approaches have operated by altering robot appearance, or by enhancing robots' communicative capabilities to allow communication that would otherwise have been impossible given their inherent morphologies. For example, Zhang et al. \cite{zhang2019evaluation} present a system to enhance human perceptions of interaction proxemics with a mixed reality robotic avatar. In this case, the physical robot is non-humanoid, but a Form Transformation VDE is utilized by overlaying a 3D AR avatar of a human above the real robot that mimics a human's gaze and body motions while moving. Through arm swinging frequency, this visualization allows the robot to effectively communicate its moving speed to nearby humans and improve subjective perceptions about the robot.

\subsubsection{Swarm Supervision} 
Finally, while many of the approaches above have focused on single robots, management of multi-robot systems is also a major challenge for robot designers and users. As more robots are added to a robotic system, the system becomes increasingly difficult to supervise and/or control. The number of robots that can be operated simultaneously is called the fan-out of a human-robot team, with robots that have high neglect tolerance and lower interaction time achieving higher fan-out \cite{olsen2004fan}. Our survey of the literature identified three papers in which AR or VR were used for this purpose. In these papers, VAM-HRI researchers have investigated how VAM technologies can increase the fan-out of robotic systems and decrease the mental load of robot operators, for example by rendering virtual imagery to display the location and status of many robots. For example, Ghiringhelli et al. \cite{ghiringhelli2014interactive} present an AR interface for swarm supervision, in which the supervisor is presented with Robot Status Visualizations VDEs (Robot Location, Robot Pose, Waypoints, Headings, and Trajectories) overlaid over each robot.


\begin{figure}
  \centering
  \includegraphics[width=\textwidth]{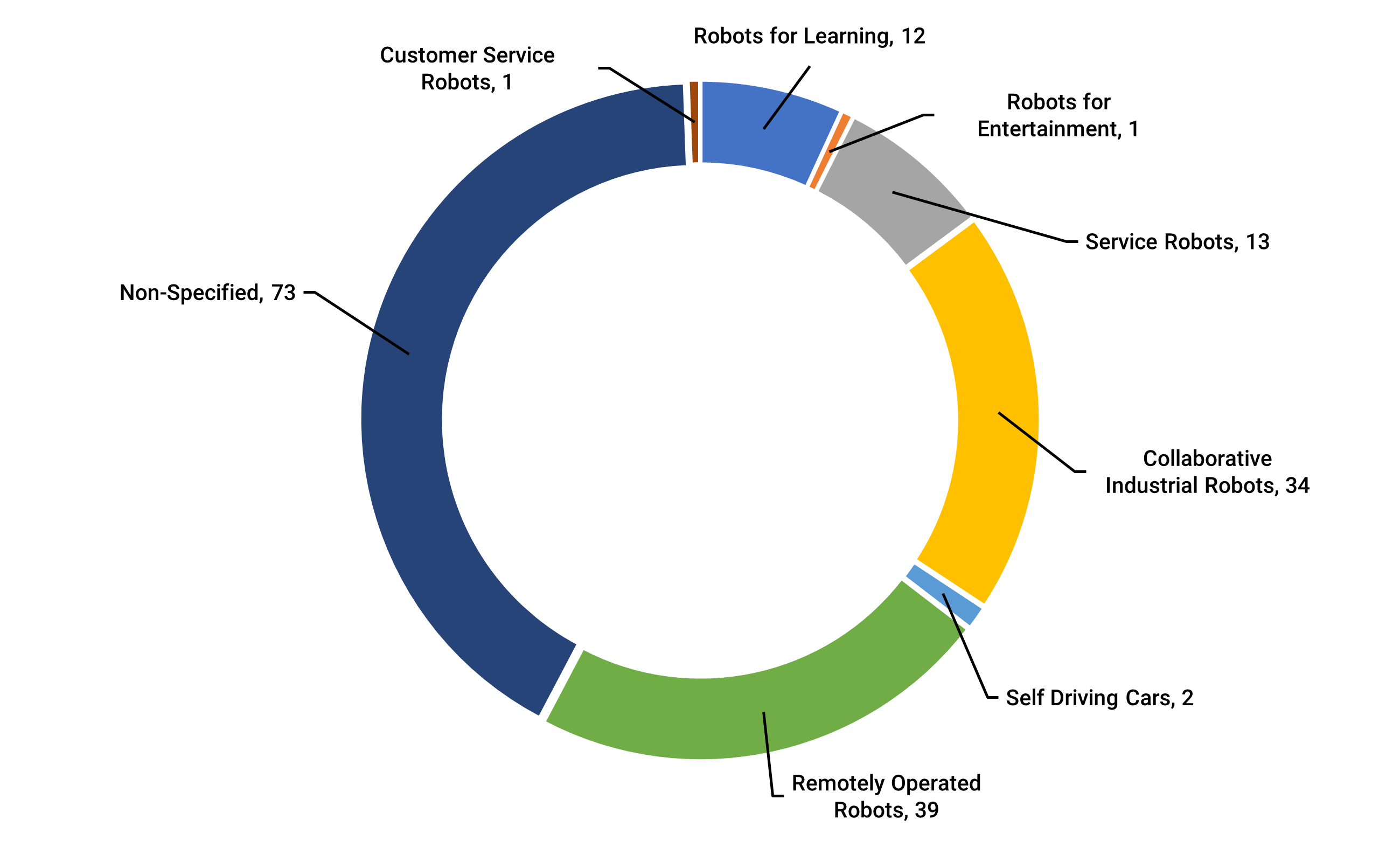}
  \caption{Application Areas}
  \label{fig:applicationArea}
\end{figure}

\subsection{Robot Domains}
While in the previous section we explored the different core robotic functionalities that VAM-HRI technologies are being used to augment, in this section we turn our attention to the high-level application domains in which such solutions are being designed and deployed. Our coverage of these domains is guided and organized according to the set of application domains delineated by Bartneck et al. in their recent textbook~\cite{bartneck2020human}: customer service robots, robots for learning, robots for entertainment, robots for healthcare and therapy, service robot, collaborative robots, self-driving cars, and remotely operated robots.

\subsubsection{Collaborative Robots} Collaborative robots, unlike traditional industrial robots, have safety features and human-friendly designs which allow human to work closely with them in manufacturing, shop floor, or maintenance contexts. Mixed and virtual reality can provide otherwise unobservable information about the robots working in these domains, such as visualization of the areas in which such robots work, the areas reachable by such robots, and the regions such robots are likely to move in the future. Making all of this information available to human teammates can make collaborative robots safer and more effective for those teammates to work with. Our survey identified 34 papers in which VR and AR are being used in these domains. 

Matsas et al. \cite{matsas2018prototyping}, for example, prototype techniques in virtual reality to improve safety in human-robot collaborative manufacturing. Specifically, in this system, a red-bordered 
circle is overlaid on human teammates' view of their environment through a virtual reality interface,  to show the limits of the robot's workspace, and a yellow wedge is drawn to show the movement range of the robot's arms. If humans enter these spaces, this wedge turns red, and a warning is displayed. 

\subsubsection{Customer Service Robots} 
Customer service robots have been used as tour guides (e.g. providing information to visitors about points of interest, and taking visitors to requested location), receptionists (e.g. providing check-in processing for hotel guests), and sales promoters (e.g. providing store promotion information  to customers). VAM technologies have not yet been applied to this domain widely: our survey identified a single paper within this domain. Specifically, Pereira et al. \cite{pereira2017augmented} demonstrated the use of an AR dialog interface that provides a compelling visual interface for customers to interact with robots in restaurant contexts. 

\subsubsection{Robots for Learning} 
Robots for learning are those operating in classroom environments, as teachers, tutors, peers, or teaching assistants -- or as the objects of study themselves -- to help make lessons more effective and engaging. With augmented reality, new knowledge can be overlaid on  physical robots so students can see this information in a spatially situated manner. Alternatively, learning content can be projected onto a surface on which students and peer robots can engage with that content together.

Our survey of prior literature identified 12 papers in this area. Johal et al. \cite{johal2019augmented}, for example,  present an educational robot that is used to teach optical concepts regarding the visible and infrared (IR) light spectrum to K-12 students during Physics classes. Johal et al. use AR in this context to visualize information about the robot's IR sensors, such as the direction of IR emitters, cone range, and intensity of the IR signal, all of which are visualized to students through Android tablets.

\subsubsection{Robots for Entertainment} Robots have also been used for entertainment, as pets, toys, exhibitions, or in the performing arts. VAM hasn't been applied widely in entertainment robotics: we identified only one paper in this domain. The robot in this paper \cite{urbani2018exploring} is not a traditional entertainment robot of the sort mentioned above but is instead a multipurpose wearable (wrist-worn) robot that has many different functionalities invoked by AR system. It consists of six interlinked servomotors fastened together using plastic brackets. This robot follows users' commands to change speeds, turn to specific angles, or set torque limits, and can wrap around or stand straight up from a user's wrist. Through an AR headset, a window appears on top of the robot, from which the users can see the robot status display, shape-changing menus, a media player on which they can watch videos, and a robot pose controller. 

\subsubsection{Robots for Healthcare and Therapy} Robots are widely applied to healthcare and therapy. They offer support for senior citizens, by helping to detect adverse medical events or by providing enhanced mobility. They are also used in the context of therapy, e.g. for people with autism spectrum disorder, people undergoing rehabilitation, and people undergoing surgery (e.g. by providing assistance in laparoscopic surgery). To support robotic surgery, AR helps to highlight anatomical structures, overlay surgery plan, robot and instrument status on the main visual source (e.g. patient body, monitor). AR can also be used to visualize the planned trajectory of the surgical robot's needle so the surgeon can check if it's valid. 

Our survey identified 71 papers in which AR and VR are used with robots in such domains. AR has been used in this domain primarily to support robotic-assisted surgery (RAS). Qian et al. \cite{qian2019augmented}, for example, present an AR system called ARssist to aid the first assistant in robotic-assisted laparoscopic surgeries. This system visualizes the robotic instruments and endoscope inside the patient body using the assistant's HMD. This approach has the potential to improve efficiency, navigation consistency and safety for instrument insertion. VAM environments have also been used in the context of therapeutic and assistive robots such as robotic wheelchairs. Zolotas et al. \cite{zolotas2019towards}, for example, present an augmented reality system that can help users control their wheelchairs safely and independently. Users of this system observe a mini-map utility that shows the wheelchair's future trajectory, potential obstacles, and potentials for collision.

\subsubsection{Service Robots} 
Service robots (as opposed to the previously discussed customer service robots) perform simple and repetitive tasks in service of humans, such as house-cleaning, delivery, or security operations, as well as other dull or dangerous tasks such as space exploration and emergency response. Many service robots work remotely by themselves. In these cases, the remote environment is displayed to the operator using 3D rendering from point cloud data or video streams. If humans and service robots are co-located, such as in certain types of search and rescue tasks, robot state information such as location, battery, condition, trajectory, and so forth, can be shown within their teammates' HMDs. Our survey of the literature found thirteen papers in this domain. 

Our survey identified 15 papers in which AR and VR are used with robots in such domains.
Martín et al., for example, \cite{san2019audio} present a multimodal AR system that allows robots to warn human teammates more effectively about hazards during navigation through unfamiliar spaces, by way of hazard area visualizations in teammates' HoloLens interfaces. Service robots can also be helpful in facilitating repetitive tasks in field domains such as agriculture. Huuskonen et al.\cite{huuskonen2019augmented}, for example, provide an AR interface that allows farmers to simultaneously monitor multiple autonomous tractors. 

\subsubsection{Self-Driving Cars}
Self-driving cars (and other autonomous vehicles) are robots that can automatically navigate between locations in large-scale human environments, especially those that can operate on semi-structured human transportation infrastructure such as roads and highways. One of the main proposed benefits of self-driving cars is to allow autonomous driving when humans are too fatigued to safely operate their vehicles. However,  current self-driving cars have a number of limitations that require humans to be able to quickly take over and intervene. Since on-road autonomous car training is dangerous and expensive, VAM technology has been applied to design autonomous car training program. During VR training, the simulators are shown in HMDs to show users information about the virtual car such as speed limit, distance traveled, and current speed. In AR training programs, the user is trained in a real car on a designated road while wearing a see-through AR HMD that can display introductory videos, car's state information, and instruction to the user. Our survey revealed two papers in this domain. For example, Sportillo et al.~\cite{sportillo2018get} present a virtual reality training program for autonomous vehicle operators, that can help train such operators to improve their ability to quickly retake vehicle control when necessary. 

\subsubsection{Remotely Operated Robots}
Remotely operated robots are robots that are controlled by humans from different places (a use case that overlaps with some of the other application domains described above). In this scenario, human operators usually receive the visualization of the remote environment in point cloud data rendering or video stream. Via VAM interface, operators can see robot status, future trajectory, heading, grasping point which are overlaid on the environment visualization. 

Our survey identified 39 papers in this domain. For example, Zollmann et al. \cite{zollmann2014flyar} present an AR interface for piloting unmanned aerial vehicles (UAV). In this paper, information about the UAV is displayed in the user's AR HMD: a virtual sphere acts as waypoint for the UAV, a virtual shadow is displayed on the ground, and a line is drawn between the waypoint and the shadow to show how high the UAV is from the ground. Many virtual spheres are connected to be UAV's trajectory, which help the user to supervise the UAV's path and intervene if they detect any potential collision. On the other hand, Kent et al. \cite{kent2017comparison} design an AR interface to help users teleoperate robots in an object manipulation task. In the AR interface, when the remote operator clicks on an object, a semi-transparent sphere appears overlaying the selected object with several blue grasping points. After the operator chooses one of the grasp points, a 3D model of robot's hand/end-effector appears for user choosing the grasping angle. After confirming the grasping pose, the real robot arm executes the grasp. 

Finally, Gharaybeh et al. \cite{chizeck2019telerobotic} present a MR system for teleoperating robot arms to defuse hazardous undetonated underwater munitions; an otherwise very dangerous task for human divers. After submerging a robot arm, its teleoperator can use visualizations of the arm's LiDAR point cloud sensor data to see the ocean floor, the undetonated munition, and other helpful visual aids, including a 3D model of the robot arm. This 3D information brings depth perception to the operators, enhancing control of the remote robot arm to defuse munitions more safely and easily.

\section*{Taxonomy Takeaways and the Future of VAM-HRI}
As seen in Figure~\ref{fig:applicationArea}, the majority of VAM-HRI papers surveyed fell into non-specified application areas, where the VAM-HRI research aimed to improve robot interactions in a more general sense by enhancing common robot functionalities (i.e., navigation, manipulation, etc.) that are required across a wide range of domains. 

However, one of the most common application areas in recent VAM-HRI research is that of collaborative industrial robots. This trend is not unexpected to see as traditional industrial robots are being replaced with this new generation of robots that are more easily deployable and are more easily able to be programmed and set up in-house. In the past traditional robots were housed behind fences due to their large size and ability to easily harm nearby humans, but collaborative robots are breaking down this physical barrier and working alongside their human counterparts. With humans now directly working with robots in industrial settings, more effort has been made to improve interactions with such robots, including the use of VAM-HRI interfaces to assist in communicating between human and robot during collaborative tasks.

As a second step in this analysis, we consider VDE use trends. While task comprehension (including sub-classes such as waypoints, trajectories, and task instructions) have historically been the most well-explored variety of VAM-HRI VDEs, other classes, such as environment comprehension VDEs, are now attracting increased attention. In contrast, object comprehension and virtual robot alteration VDEs are the newest variants to be explored, first introduced in 2004 and 2006 respectively. These VDEs are still relatively unexplored and may be fruitful avenues for further research. 

Overall, in this paper we have explored a wide range of research that demonstrates that VAM-HRI is an active and rapidly growing research area. We believe, however, that this field is currently hindered by a lack of precise terminology and theoretical models that explain how work in the field may connect and build off each other. The taxonomy we have presented for identifying, grouping, and classifying key design elements across VAM-HRI systems helps to address this issue and highlights potential design elements that have yet to appear in the research literature, which may serve as fertile ground for future research. For example, in the creation of this taxonomy it was realized that Special Effect Alteration VDEs had yet to be explored in scientific literature; however, when viewing the VAM-HRI work done to date from the high-level view of the VDE Table, this hole in the research landscape is made clear. It is our hope that other researchers are able to use the table in a similar fashion to guide their own efforts, and that our work will help VAM-HRI grow into a mainstream field by providing researchers in the community with the necessary lexicon for easily understanding, describing, and referencing the designs in their own systems with other relevant work being performed, while also helping researchers reason about what areas require further exploration or represent entirely novel areas of inquiry. 

Additionally, the VAM VDE Taxonomy can be used as a catalogue and/or cookbook for robot designers interested in enhancing their robots and their interactions through VAM technology. We envision developers being able see all the VDEs available, including their hierarchical categories and classes, when deploying VAM-HRI systems and being able to pick and choose the VDE(s) that best addresses their VAM-HRI design's challenges and purpose. 

Finally, it is important to note that as research and technology in this emerging field becomes more mature, the taxonomy presented in this article is destined to change and grow as time goes on. This taxonomy thus serves as a jumping off point for organizing and inspiring the future work performed within this field. 

\bibliographystyle{ACM-Reference-Format}
\bibliography{sample-base}










\end{document}